\documentclass{article}
\usepackage{spconf,amsmath,graphicx}
\usepackage[hidelinks]{hyperref}
\usepackage{booktabs} 
\usepackage{tikz}
\usepackage{pgfplots}
\pgfplotsset{compat=1.3}
\usepackage{subcaption}
\usepackage{xcolor}
\usepackage{xspace}

\usepackage{amsfonts}

\newcommand{\norm}[1]{\lVert #1 \rVert} 

\newcommand{\Lc}{\mathcal{L}}

\newcommand{\Uc}{\mathcal{U}}

\newcommand{\Eb}{\mathbb{E}}

\newcommand{\Rb}{\mathbb{R}}

\newcommand{\uv}{\mathbf{u}}
\newcommand{\vv}{\mathbf{v}}

\newcommand{\xv}{\mathbf{x}}
\newcommand{\yv}{\mathbf{y}}
\newcommand{\zv}{\mathbf{z}}

\newcommand{\Wv}{\mathbf{W}}
\newcommand{\Xv}{\mathbf{X}}

\newcommand{\alphav     }{\boldsymbol \alpha     }

\newcommand{\thetav     }{\boldsymbol \theta     }

\newcommand{\lambdav    }{\boldsymbol \lambda    }

\newcommand{\methodfull}{Heterogeneous Joint Pruning\xspace}
\newcommand{\method}{HJ-Pruning\xspace}

\newcommand{\methodOSfull}{\method based on the overall model size\xspace}
\newcommand{\methodSSfull}{\method based on separate model sizes\xspace}
\newcommand{\methodOMfull}{\method based on the overall MACs\xspace}

\newcommand{\methodOS}{\method-Size\xspace}
\newcommand{\methodSS}{\method-SepSize\xspace}
\newcommand{\methodOM}{\method-MAC\xspace}

\title{Structured Pruning of Self-Supervised Pre-trained Models\\
for Speech Recognition and Understanding}

\name{Yifan Peng$^2$\sthanks{Work done during an internship at ASAPP.}, Kwangyoun Kim$^1$, Felix Wu$^1$, Prashant Sridhar$^1$, Shinji Watanabe$^2$}
\address{
$^1$ASAPP Inc., Mountain View, CA, USA\quad
$^2$Carnegie Mellon University, Pittsburgh, PA, USA\\
\small{\texttt{\{yifanpen,swatanab\}@andrew.cmu.edu}\quad \texttt{\{kkim,fwu,psridhar\}@asapp.com}}
}

\begin{document}
\ninept
\maketitle
\begin{abstract}
Self-supervised speech representation learning (SSL) has shown to be effective in various downstream tasks, but SSL models are usually large and slow. Model compression techniques such as pruning aim to reduce the model size and computation without degradation in accuracy. Prior studies focus on the pruning of Transformers; however, speech models not only utilize a stack of Transformer blocks, but also combine a frontend network based on multiple convolutional layers for low-level feature representation learning. This frontend has a small size but a heavy computational cost. In this work, we propose three task-specific structured pruning methods to deal with such heterogeneous networks. Experiments on LibriSpeech and SLURP show that the proposed method is more accurate than the original wav2vec2-base with 10\% to 30\% less computation, and is able to reduce the computation by 40\% to 50\% without any degradation.
\end{abstract}

\section{Introduction}
\label{sec:intro}

Self-supervised speech representation learning (SSL) has achieved great success in a variety of speech processing tasks~\cite{superb, ssl-review, ssl-for-asr, ssl-for-se-ss, slue, espnet-slu, ssl-for-slu}. However, SSL pre-trained models (e.g., wav2vec2~\cite{wav2vec2}, HuBERT~\cite{hubert} and WavLM~\cite{wavlm}) usually require large memory and high computational cost. Hence, it is difficult to deploy them in real-world applications. Recent studies have utilized model compression techniques to reduce the model size and computation without degradation in accuracy. One common approach is knowledge distillation~\cite{hinton2015distilling}, which trains a small student model with a pre-specified architecture to match the soft targets generated by a large pre-trained model. Distillation has shown to be effective in natural language processing (NLP)~\cite{distilbert, tinybert} and speech processing~\cite{shrinking-bigfoot, distilhubert, lighthubert, fithubert}, but it usually performs general distillation using large amounts of unlabeled data before task-specific distillation or fine-tuning. This can make the training procedure computationally expensive.

Another compression technique is pruning, which extracts a compact and accurate subnetwork from the original model. Pruning has been widely used in computer vision (CV)~\cite{han2015pruning, filter-pruning-conv, network-slimming, platon-icml22} and NLP~\cite{platon-icml22, asapp-pruning, cofi, super-tickets-nlp}. For speech models, \cite{structured-prune-rnn, prune-lstm} prune recurrent neural networks (RNNs) for resource-limited applications.
Another work~\cite{deliang-compressing-enhancement} prunes deep neural networks (DNNs) based speech enhancement models using the sparse group lasso regularization~\cite{group-sparse-reg}. These studies do not consider SSL models.
PARP~\cite{parp} is one of the first pruning methods designed for SSL speech models, which prunes individual weights based on magnitudes. Despite its good performance in low-resource automatic speech recognition (ASR), PARP is a type of unstructured pruning and thus cannot achieve an actual speedup without an efficient sparse matrix computation library, which is not usually available in many deployment scenarios. 
Another limitation is that PARP only prunes the Transformer layers while keeping the convolutional feature extractor (CNN) fixed. As discussed in~\cite{felix-sew}, although the CNN has much fewer parameters than the Transformer, its computational cost is large and cannot simply be ignored. For example, in wav2vec2-base, the CNN has less than 5\% of the total parameters but its computational cost is nearly 33\% in terms of multiply-accumulate (MAC) operations for a 10-second audio.

In this work, we propose \method (\methodfull) where both CNN and Transformer components are pruned jointly.
We consider three variants:
(a) \textit{\methodOSfull} sets a single sparsity for the entire model size.
(b) \textit{\methodSSfull} introduces a separate sparsity hyperparameter for each component which allows fine-grained tuning to find a trade-off between CNN and Transformer.
(c) \textit{\methodOMfull} uses multiply–accumulate (MAC) operations as the sparsity criterion to find the best allocation of the computation budget across different components.
We evaluate our methods in the ASR and spoken language understanding (SLU) tasks. Experiments on LibriSpeech and SLURP show that all \method methods outperform the previous Transformer-only pruning strategy. Our \methodOM is more accurate than the original wav2vec2-base with 10\% to 30\% less computation, and is able to reduce the computation by 40\% to 50\% without any degradation.

\section{Background}

\subsection{Self-supervised pre-trained speech models}
\label{subsec:ssl-models}
We evaluate the pruning algorithms mainly using wav2vec2~\cite{wav2vec2}, but our proposed methods can be easily applied to other SSL models with as a similar architecture such as HuBERT~\cite{hubert} (see Sec.~\ref{subsec:other-compression}), SEW-D~\cite{felix-sew}, and WavLM~\cite{wavlm}. The wav2vec2-base model (pretrained on Librispeech 960h~\cite{librispeech-corpus}) consists of a convolutional feature extractor (CNN) and a Transformer~\cite{transformer} encoder. The CNN contains seven temporal convolutions with 512 channels and GeLU~\cite{gelu} activations. The Transformer encoder is a stack of 12 Transformer layers with a hidden dimension of 768 and 12 attention heads.

\subsection{Structured pruning using $L_0$ regularization}
\label{subsec:pruning-l0reg}

We follow \cite{asapp-pruning, cofi, l0sparse-iclr18} to formulate the structured pruning task as a regularized learning problem, which aims to learn a sparse model. Let $f(\cdot;\thetav)$ be a model with parameters $\thetav = \{\theta_j\}_{j=1}^n$, where each $\theta_j$ is a group of parameters (e.g., an attention head) and $n$ is the number of groups. The pruning decisions are given by a set of binary variables called \textit{gates}: $\zv = \{z_j\}_{j=1}^n$ where $z_j\in \{0, 1\}$. The model parameters after pruning are $ \tilde{\thetav} = \{\tilde{\theta}_j\}_{j=1}^n$ such that $\tilde{\theta}_j = \theta_j z_j$. We usually sample gates from some distributions (e.g., Bernoulli) and update their parameters during training. Suppose the gates follow a distribution $q(\zv;\alphav)$ with parameters $\alphav = \{\alpha_j\}_{j=1}^n$, then our training objective is:
\begin{align}
\label{eq:total-loss}
    \min_{\thetav, \alphav} ~~
    \Eb_{\zv \sim q}\left[\frac{1}{D} \sum_{i=1}^D \Lc(f(\xv_i;\tilde{\thetav}),\yv_i) + \lambda \norm{\tilde{\thetav}}_0 \right],
\end{align}
where $\{(\xv_i, \yv_i)\}_{i=1}^D$ is the training data containing $D$ samples, $\Lc$ is the training loss (i.e., CTC loss for ASR, cross entropy loss for SLU), and $\lambda > 0$ is a hyperparameter to control the sparsity. However, it is intractable to optimize Eq.~\eqref{eq:total-loss} using gradient descent because the gates are discrete. Louizos et al.~\cite{l0sparse-iclr18} propose a reparameterization trick to make the loss differentiable, which has been widely used in sparse model learning. Here we only introduce their final approach. Please refer to \cite{l0sparse-iclr18} for the derivation. Louizos et al. adopt the Hard Concrete Distribution~\cite{l0sparse-iclr18} to model the gates $\zv$:
\begin{align}
\label{eq:hard-concrete}
\begin{split}
    \uv &\sim \Uc(0,1), ~~
    \vv(\alphav) = \sigma\left(\frac{1}{\beta}\left(\log\frac{\uv}{1-\uv} + \log\alphav\right)\right), \\
    \bar{\vv}(\alphav) &= (r-l)\cdot\vv(\alphav) + l, ~~ \zv = \min(1,\max(0,\bar{\vv}(\alphav))),
\end{split}
\end{align}
where $\Uc(0,1)$ is a uniform distribution over the interval $[0,1]$, $\sigma(\cdot)$ is the sigmoid function and $\beta$ is a temperature constant. The actual parameters are $\alphav$. $l < 0$ and $r > 0$ are two constants to stretch the output of sigmoid to $[l, r]$, which is finally rectified to $[0,1]$. It is proven that the first term in Eq.~\eqref{eq:total-loss} now becomes differentiable w.r.t. all parameters. We can write the second term in a closed-form expression based on the distribution of $\zv$ shown in Eq.~\eqref{eq:hard-concrete}:
\begin{equation}
\label{eq:expected-norm}
    \Eb_{\zv}\left[\norm{\tilde{\thetav}}_0\right]
    = \sum_{j=1}^n P(z_j \neq 0)
    = \sum_{j=1}^n \sigma\left( \log \alpha_j - \beta \log\frac{-l}{r} \right),
\end{equation}
which is also differentiable. $P(\cdot)$ denotes the probability.

Now we can train a sparse model using Eq.~\eqref{eq:total-loss}. However, it is difficult to exactly control the pruned model size~\cite{asapp-pruning, cofi}. Instead of adding a regularizer $\lambda \norm{\tilde{\thetav}}_0$, prior studies~\cite{asapp-pruning, cofi} suggest optimizing the training loss subject to an explicit equality constraint on sparsity:
\begin{align}
    \label{eq:min-st-cons}
    \min_{\thetav, \alphav} ~~
    \Eb_{\zv}\left[\frac{1}{D} \sum_{i=1}^D \Lc(f(\xv_i;\tilde{\thetav}),\yv_i) \right]~~\text{s.t.}~~s(\alphav)=t,
\end{align}
where $s(\alphav)$ is the current sparsity and $t$ is a pre-specified target sparsity. The sparsity is defined as the percentage of parameters that are pruned. Similar to Eq.~\eqref{eq:expected-norm}, given the current parameters $\alphav$, we can calculate the expected number of nonzero gates in every module of the model. Recall that each gate is associated with a group of parameters. Hence, we know the expected number of parameters that are still kept, which further gives us the sparsity $s(\alphav)$. Eq.~\eqref{eq:min-st-cons} can be rewritten as an adversarial game according to the augmented Lagrangian method~\cite{asapp-pruning}:
\begin{align}
\label{eq:minimax}
    \max_{\lambdav} \min_{\thetav, \alphav} ~~ \Eb_{\zv} \left[\frac{1}{D} \sum_{i=1}^D \Lc(f(\xv_i;\tilde{\thetav}),\yv_i) \right] + g(\lambdav, \alphav),\\
\label{eq:lagterm}
    g(\lambdav, \alphav) = \lambda_1 (s(\alphav)-t) + \lambda_2 (s(\alphav) - t)^2,
\end{align}
where $\lambda_1, \lambda_2 \in \Rb$ are two Lagrange multipliers that are jointly trained with other parameters.
Once the game reaches equilibrium, the equality constraint will be satisfied. Hence, we can precisely control the sparsity of the pruned model. To facilitate training, we linearly increase the target sparsity $t$ from zero to the desired value.

\subsection{Structured pruning of Transformer layers}
\label{subsec:prune-transformer}

A Transformer~\cite{transformer} layer consists of a multi-head self-attention (MHA) block and a position-wise feed-forward network (FFN). We consider three pruning units, i.e., attention heads (12 per layer), intermediate size of FFN (3072 per layer), and the model's hidden size (768).
We define a gate for each pruning unit. Given an input sequence $\Xv\in \Rb^{T\times d}$ of length $T$ and feature size $d$, the MHA and FFN at a particular layer are the following:
\begin{align}
    \mathrm{MHA}(\Xv) &= \sum_{k=1}^{h} (z^\text{head}_k \cdot \text{ATT}(\Xv; \Wv^\text{att}_k) ), \\
    \text{FFN}(\Xv) &= \text{GeLU}(\Xv\Wv^\text{ffn}_1)\cdot \text{diag}(\zv^\text{int})\cdot \Wv^\text{ffn}_2,
\end{align}
where $\text{ATT}(\cdot; \Wv^\text{att}_k)$ denotes the $k$-th attention head parameterized by $\Wv^\text{att}_k$, and $z_k^\text{head}$ is a scalar gate. There are $h$ heads in total. $\zv^\text{int}$ is a $d^\text{int}$-dimensional gate for the FFN intermediate size. $\text{diag}(\cdot)$ creates a diagonal matrix with its argument vector on the diagonal. GeLU is an activation~\cite{gelu}. FFN has two linear layers $\Wv^\text{ffn}_1 \in \Rb^{d\times d^\text{int}}, \Wv^\text{ffn}_2 \in \Rb^{d^\text{int}\times d}$. Each Transformer layer has its own gates and their parameters are independent. For the main hidden size, we define a gate $\zv^\text{hid}$ of size $d$ and share it across layers as in~\cite{cofi}.

\section{Proposed Methods}

\subsection{Joint pruning based on the model size}
\label{subsec:joint-size}

As introduced in Sec.~\ref{sec:intro}, the convolutional feature extractor (CNN) in SSL models is small in size but heavy in computation. To optimize the overall computation, we propose to jointly prune the CNN and Transformer. We have introduced the pruning units for Transformer in Sec.~\ref{subsec:prune-transformer}. For CNN, we prune convolution channels by introducing gate variables for every channel in every CNN layer, i.e., each output channel is multiplied with a gate. To train the model using Eq.~\eqref{eq:minimax}, we need to define the model sparsity $s(\alphav)$. Our first proposed method is \textbf{\methodOS} (\methodOSfull), which can be viewed as a direct extension from prior work~\cite{asapp-pruning, cofi}. Specifically, given the current distribution parameters $\alphav$, we can calculate the probability of each gate being nonzero (i.e., the corresponding module is kept) as in Eq.~\eqref{eq:expected-norm}. We then know the current sizes of all modules, including the model's hidden size, CNN channels, attention heads, and FFN intermediate sizes. Based on these sizes, we can compute the percentage of parameters that are pruned, which is the overall size sparsity $s^\text{all}_\text{size}(\alphav)$.

However, Sec.~\ref{subsec:main-results} shows that this approach does not work well in practice, because the CNN has much fewer parameters than the Transformer. If we simply set an overall sparsity, parameters will be pruned mainly from Transformer. To solve this problem, we propose the second method, i.e., \textbf{\methodSS} (\methodSSfull). We calculate the size sparsity separately for CNN ($s^\text{cnn}_\text{size}(\alphav)$) and Transformer ($s^\text{trans}_\text{size}(\alphav)$). We also specify separate target sparsities $t^\text{cnn}_\text{size}, t^\text{trans}_\text{size}$ and extend Eq.~\eqref{eq:lagterm} to have two terms:
\begin{align}
\label{eq:prune-sep-size}
\begin{split}
    g_\text{size} &= \lambda_1^\text{cnn} (s^\text{cnn}_\text{size}(\alphav) - t^\text{cnn}_\text{size}) + \lambda_2^\text{cnn} (s^\text{cnn}_\text{size}(\alphav) - t^\text{cnn}_\text{size})^2 \\
    &+ \lambda_1^\text{trans} (s^\text{trans}_\text{size}(\alphav) - t^\text{trans}_\text{size}) + \lambda_2^\text{trans} (s^\text{trans}_\text{size}(\alphav) - t^\text{trans}_\text{size})^2.
\end{split}
\end{align}
As shown in Sec.~\ref{subsec:main-results}, this method achieves strong performance. However, it requires careful tuning of the separate target sparsities. We always need to search over the two sparsities to meet a particular budget, which is computationally expensive.

\subsection{Joint pruning based on the MACs}
\label{subsec:prune-macs}

The third method we propose is \textbf{\methodOM} (\methodOMfull). Unlike prior methods, we prune the entire model to directly meet a computational budget measured by MACs. We follow the formulas used in the DeepSpeed flops profiler to calculate MACs.~\footnote{\url{https://github.com/microsoft/DeepSpeed}}
For an input sequence of length $T$ and hidden size $d$, the MACs for each MHA and FFN block are as follows:
\begin{align}
    \text{MAC}^\text{mha} &= 4Thdd^\text{head} + 2T^2hd^\text{head},\\
    \text{MAC}^\text{ffn} &= 2Tdd^\text{int},
\end{align}
where $h$ is the number of attention heads and $d^\text{head}$ is the size per head. $d^\text{int}$ denotes the intermediate size of FFN.
The MACs of a 1-D convolution can be computed by
\begin{align}
    \text{MAC}^\text{cnn} = T^\text{out} C^\text{out} C^\text{in} K,
\end{align}
where $T^\text{out}$ is the output length and $K$ is the kernel size. $C^\text{in}$ and $C^\text{out}$ are the input and output channels, respectively. Note that $h, d, d^\text{int}, C^\text{in}, C^\text{out}$ are calculated from the current gate distributions (similar to Eq.~\eqref{eq:expected-norm}). They are differentiable functions of $\alphav$. 
We define the percentage of MACs that are pruned as the MACs-based sparsity $s^\text{all}_\text{macs}(\alphav)$.~\footnote{The computation of MACs also depends on the sequence length $T$, because MHA has quadratic complexity w.r.t. $T$. We use 10 seconds to compute MACs in our experiments. This is a ``virtual'' length used only for computing MACs. We do not modify any training utterances.} It is differentiable w.r.t. parameters $\alphav$. Hence, we can still train the model using Eq.~\eqref{eq:minimax}.

\section{Experiments}

\subsection{Experimental setup}

We focus on task-specific structured pruning of SSL speech models. We mainly prune wav2vec2-base, but we also show that our methods can be directly applied to HuBERT-base in Sec.~\ref{subsec:other-compression}. We conduct experiments using PyTorch~\cite{pytorch} and HuggingFace's transformers~\cite{huggingface-transformers}.  Our implementation of the basic pruning algorithm is based on prior work in NLP~\cite{cofi}. For each task, we add a linear layer on top of the pre-trained SSL model and fine-tune the entire model to obtain an unpruned model. Then, we prune this fine-tuned model to reach a specific sparsity using Eq.~\eqref{eq:minimax}. We employ an AdamW optimizer and a linear learning rate scheduler for all experiments.

\textbf{ASR}: The 100-hour clean set of LibriSpeech~\cite{librispeech-corpus} is utilized. In Sec.~\ref{subsec:robustness}, the Tedlium~\cite{tedlium-corpus} test set is used as out-of-domain data to demonstrate the robustness of structured pruning. The training loss is CTC~\cite{ctc}. We fine-tune a pre-trained model for 25 epochs and prune for 30 epochs with a learning rate of 1.5e-4 and a batch size of 64. The target sparsity is linearly increased to the desired value during the first 5 epochs. The learning rate of $\alphav$ and $\lambdav$ is selected from $\{0.02, 0.05\}$. The pruned model is fine-tuned for another 10 epochs with a learning rate of 5e-5. The learning rate warmup steps are 3k, 3k, and 1k for training, pruning, and fine-tuning, respectively. 

\textbf{SLU}: The SLURP corpus~\cite{slurp-corpus} is used for intent classification. A pre-trained SSL model is fine-tuned for 50 epochs and pruned for 50 epochs with a learning rate of 1e-4 and a batch size of 80. The final fine-tuning has 10 epochs with a learning rate of 1e-5. The learning rate warmup is performed for 4k, 4k, 1k steps for training, pruning, and fine-tuning, respectively. Other configs are the same as ASR.

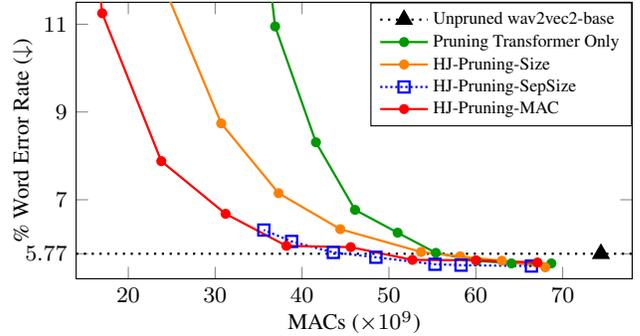
\begin{figure}[t]
\centering
\resizebox {0.99\linewidth} {!} {
\begin{tikzpicture}
	\begin{axis}[
		xlabel=MACs ($\times 10^9$),
		ylabel=\% Word Error Rate ($\downarrow$),
		xtick={0, 10,...,100},
            ytick={5.77, 7,9,11,13},
    	xmin=14,
            xmax=78,
            ymin=5.2,
            ymax=11.5,
            ylabel shift = -9pt,
            xlabel shift = -3pt,
            label style={font=\footnotesize},
            ticklabel style={font=\footnotesize},
            width=\linewidth,
            height=0.59\linewidth,
	    every axis plot/.append style={thick},
            legend cell align={left},
            legend columns=1,
            legend style={at={(1,1)},anchor=north east,nodes={scale=0.7, transform shape}}
		]

        \addplot[color=black, dotted, mark=triangle*, mark options={scale=1.5, solid, fill=black}] coordinates {
(200, 5.77)
(74.4, 5.77)
(0, 5.77)
	};
        \addlegendentry{Unpruned wav2vec2-base};
        
	\addplot[color=green!60!black, solid, mark=*, mark options={scale=0.7, solid}] coordinates {
(68.7, 5.55)
(64.1, 5.55)
(60.1, 5.61)
(55.4, 5.79)
(51.0, 6.25)
(46.1, 6.77)
(41.6, 8.31)
(36.9, 10.95)
(32.3, 18.70)
	}; 
        \addlegendentry{Pruning Transformer Only};
        
	\addplot[color=orange, mark=*, solid, mark options={scale=0.7, solid}] coordinates {
(68.0, 5.46)
(63.0, 5.61)
(58.2, 5.71)
(53.7, 5.81)
(44.4, 6.33)
(37.3, 7.15)
(30.7, 8.74)
(24.0, 11.91)
(16.1, 22.94)
	};
        \addlegendentry{\methodOS};

        \addplot[color=blue, mark=square, densely dotted, mark options={scale=1, solid}] coordinates {
(66.4, 5.49)
(58.3, 5.51)
(55.3, 5.53)
(48.5, 5.69)
(43.6, 5.80)
(38.8, 6.05)
(35.6, 6.31)
	};
        \addlegendentry{\methodSS};
 
        \addplot[color=red, mark=*, solid, mark options={scale=0.7, solid, fill=red}] coordinates {
(67.1, 5.57)
(60.0, 5.62)
(52.7, 5.63)
(45.6, 5.92)
(38.2, 5.95)
(31.2, 6.68)
(23.8, 7.88)
(17.0, 11.25)
(9.9, 24.07)
	};
        \addlegendentry{\methodOM};
	\end{axis}
\end{tikzpicture}
} 
    \vskip -0.15in
  \caption{Word Error Rate (\%) vs. Multiply-Accumulate Operations on LibriSpeech test-clean. Our proposed \method methods consistently outperform the baseline.
  }
  \label{fig:asr-main}
  \vskip -0.1in
\end{figure}
\begin{figure}[t]
\centering
\resizebox {0.98\linewidth} {!} {
\begin{tikzpicture}
	\begin{axis}[
		xlabel=MACs ($\times 10^9$),
		ylabel=\% Accuracy ($\uparrow$),
		xtick={0, 10,...,100},
            ytick={75,80,85,86.1},
    	xmin=8,
            xmax=78,
            ymin=75,
            ymax=87.5,
            ylabel shift = -9pt,
            xlabel shift = -3pt,
            label style={font=\footnotesize},
            ticklabel style={font=\footnotesize},
            width=\linewidth,
            height=0.6\linewidth,
	    every axis plot/.append style={thick},
            legend cell align={left},
            legend columns=1,
            legend style={at={(1,0.05)},anchor=south east,nodes={scale=0.7, transform shape}}
		]
  
        \addplot[color=black, dotted, mark=triangle*, mark options={scale=1.5, solid, fill=black}] coordinates {
(0, 86.1)
(74.5, 86.1)
(200, 86.1)
	};
        \addlegendentry{ Unpruned wav2vec2-base};
        
	\addplot[color=green!60!black, solid, mark=*, mark options={scale=0.7, solid}] coordinates {
(70.2, 86.9)
(65.3, 86.8)
(60.7, 86.5)
(56.2, 86.6)
(51.4, 86.5)
(46.6, 85.9)
(41.8, 85.1)
(37.3, 83.1)
(32.6, 78.5)
	}; 
        \addlegendentry{Pruning Transformer Only};
        
	\addplot[color=orange, mark=*, solid, mark options={scale=0.7, solid}] coordinates {
(67.1, 86.9)
(61.3, 86.7)
(56.6, 86.8)
(50.5, 86.6)
(45.2, 86.7)
(39.0, 86.2)
(32.0, 85.2)
(24.1, 82.8)
(13.9, 76.7)
	};
        \addlegendentry{\methodOS};

        \addplot[color=blue, mark=square, densely dotted, mark options={scale=1, solid}] coordinates {
(67.5, 87.0)
(61.9, 87.0)
(57.1, 87.1)
(46.9, 86.9)
(42.3, 86.8)
(33.1, 86.0)
(27.9, 85.4)
(26.0, 85.1)
	};
        \addlegendentry{\methodSS};
        
        \addplot[color=red, mark=*, solid, mark options={scale=0.7, solid, fill=red}] coordinates {
(67.3, 86.6)
(60.4, 86.6)
(53.2, 86.5)
(45.9, 86.2)
(38.7, 86.5)
(35.2, 86.5)
(31.6, 85.6)
(24.6, 85.0)
(17.1, 82.8)
(9.8, 75.8)	
        };
        \addlegendentry{\methodOM};
    
	\end{axis}
\end{tikzpicture}
}
    \vskip -0.15in
  \caption{Intent Classification Accuracy (\%) vs. Multiply-Accumulate Operations on the SLURP test set. Our proposed \method methods consistently outperform the baseline.}
  \label{fig:slu-main}
  \vskip -0.2in
\end{figure}
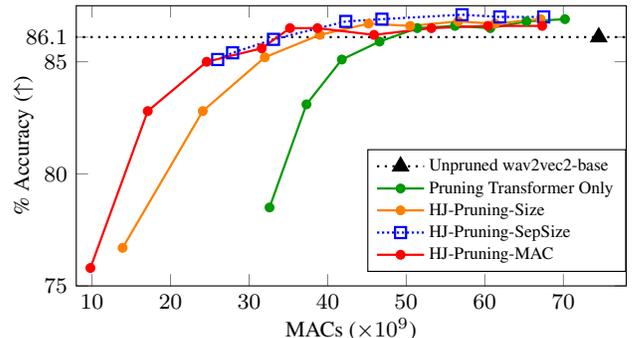
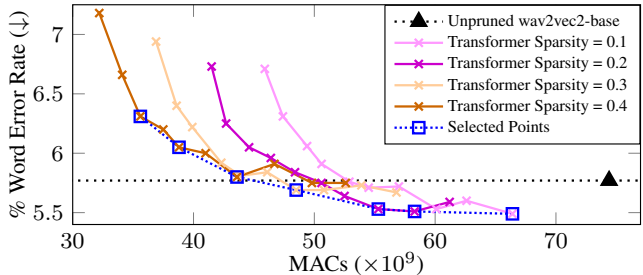
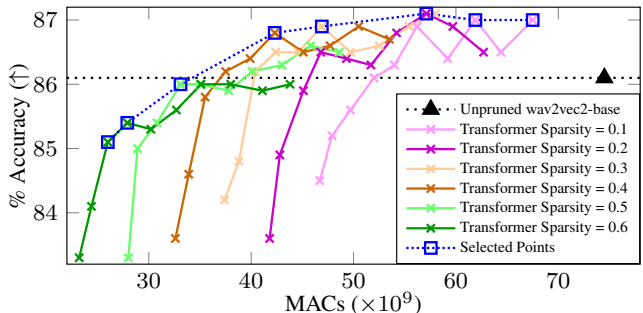
\begin{figure}
     \begin{subfigure}[b]{\linewidth}
          \centering
          \resizebox{\linewidth}{!}{\begin{tikzpicture}
	\begin{axis}[
		xlabel=MACs ($\times 10^9$),
		ylabel=\% Word Error Rate ($\downarrow$),
		xtick={0, 10,...,100},
    	xmin=30,
            xmax=77,
            ymin=5.4,
            ymax=7.25,
            ylabel shift = -5pt,
            xlabel shift = -5pt,
            label style={font=\footnotesize},
            ticklabel style={font=\footnotesize},
            height = 0.5\linewidth,
            width = \linewidth,
	    every axis plot/.append style={thick},
            legend cell align={left},
            legend columns=1,
            legend style={at={(1,1)},anchor=north east,nodes={scale=0.65, transform shape}}
		]
  
        \addplot[color=black, dotted, mark=triangle*, mark options={scale=1.5, solid, fill=black}] coordinates {
(200, 5.77)
(74.4, 5.77)
(0, 5.77)
	};
        \addlegendentry{ Unpruned wav2vec2-base};
        
	\addplot[color={rgb,255:red,255;green,153;blue,255}, mark=x, solid, mark options={scale=1, solid}] coordinates {
(66.4, 5.49)
(62.6, 5.60)
(60.1, 5.53)
(57.0, 5.72)
(54.5, 5.71)
(52.9, 5.76)
(50.6, 5.91)
(49.4, 6.06)
(47.4, 6.31)
(45.9, 6.71)
	};
        \addlegendentry{ Transformer Sparsity = 0.1};
        
        \addplot[color={rgb,255:red,204;green,0;blue,204}, mark=x, solid, mark options={scale=1, solid}] coordinates {
(61.2, 5.59)
(58.3, 5.51)
(55.3, 5.53)
(52.5, 5.64)
(50.6, 5.75)
(48.4, 5.84)
(46.4, 5.96)
(44.6, 6.05)
(42.7, 6.25)
(41.5, 6.73)
	};
        \addlegendentry{ Transformer Sparsity = 0.2};
        
        \addplot[color={rgb,255:red,255;green,204;blue,153}, mark=x, solid, mark options={scale=1, solid}] coordinates {
(56.8, 5.67)
(53.9, 5.73)
(50.8, 5.69)
(48.5, 5.69)
(46.1, 5.84)
(43.9, 5.81)
(42.3, 5.92)
(39.9, 6.22)
(38.6, 6.40)
(36.9, 6.94)
	};
        \addlegendentry{ Transformer Sparsity = 0.3};
 
        \addplot[color={rgb,255:red,204;green,102;blue,0}, mark=x, solid, mark options={scale=1, solid}] coordinates {
(52.6, 5.75)
(49.8, 5.75)
(46.7, 5.91)
(43.6, 5.80)
(41.0, 6.00)
(38.8, 6.05)
(37.5, 6.20)
(35.6, 6.31)
(34.1, 6.66)
(32.2, 7.18)
	};
        \addlegendentry{ Transformer Sparsity = 0.4 };

        \addplot[color=blue, mark=square, densely dotted, mark options={scale=1, solid}] coordinates {
(66.4, 5.49)
(58.3, 5.51)
(55.3, 5.53)
(48.5, 5.69)
(43.6, 5.80)
(38.8, 6.05)
(35.6, 6.31)
	};
        \addlegendentry{Selected Points};

	\end{axis}
\end{tikzpicture}}
          \vskip -0.1in
          \caption{Word Error Rates (\%) on LibriSpeech test-clean.}
          \label{fig:asr-sep-size}
     \end{subfigure}
     
     \vskip 0.05in
     \begin{subfigure}[b]{\linewidth}
          \centering
          \resizebox{\linewidth}{!}{\begin{tikzpicture}
	\begin{axis}[
		xlabel=MACs ($\times 10^9$),
		ylabel=\% Accuracy ($\uparrow$),
		xtick={0, 10,...,100},
    	xmin=22,
            xmax=78,
            ymin=83.2,
            ymax=87.2,
            width = \linewidth,
            height = 0.55\linewidth,
            ylabel shift = -5pt,
            xlabel shift = -5pt,
            label style={font=\footnotesize},
            ticklabel style={font=\footnotesize},
	    every axis plot/.append style={thick},
            legend cell align={left},
            legend columns=1,
            legend style={at={(1,0)},anchor=south east,nodes={scale=0.6, transform shape}}
		]
  
        \addplot[color=black, dotted, mark=triangle*, mark options={scale=1.5, solid, fill=black}] coordinates {
(0, 86.1)
(74.5, 86.1)
(200, 86.1)
	};
        \addlegendentry{ Unpruned wav2vec2-base};
        
	\addplot[color={rgb,255:red,255;green,153;blue,255},mark=x, solid, mark options={scale=1, solid}] coordinates {
(67.5, 87.0)
(64.4, 86.5)
(61.9, 87.0)
(59.2, 86.4)
(56.2, 86.9)
(54.0, 86.3)
(52.0, 86.1)
(49.7, 85.6)
(47.9, 85.2)
(46.7, 84.5)
	};
        \addlegendentry{Transformer Sparsity = 0.1};
        
        \addplot[color={rgb,255:red,204;green,0;blue,204},mark=x, solid, mark options={scale=1, solid}] coordinates {
(62.7, 86.5)
(59.7, 86.9)
(57.1, 87.1)
(54.2, 86.8)
(51.7, 86.3)
(49.3, 86.4)
(46.8, 86.5)
(45.1, 85.9)
(42.8, 84.9)
(41.8, 83.6)
	};
        \addlegendentry{Transformer Sparsity = 0.2};
        
        \addplot[color={rgb,255:red,255;green,204;blue,153},mark=x, solid, mark options={scale=1, solid}] coordinates {
(58.0, 87.1)
(55.4, 86.9)
(52.5, 86.6)
(49.7, 86.5)
(46.9, 86.9)
(44.6, 86.5)
(42.4, 86.5)
(40.4, 86.2)
(38.8, 84.8)
(37.4, 84.2)
	};
        \addlegendentry{Transformer Sparsity = 0.3};
 
        \addplot[color={rgb,255:red,204;green,102;blue,0},mark=x, solid, mark options={scale=1, solid}] coordinates {
(53.5, 86.7)
(50.5, 86.9)
(47.7, 86.6)
(45.1, 86.5)
(42.3, 86.8)
(39.9, 86.4)
(37.5, 86.2)
(35.5, 85.8)
(33.9, 84.6)
(32.6, 83.6)
	};
        \addlegendentry{Transformer Sparsity = 0.4};

        \addplot[color={rgb,255:red,102;green,255;blue,102},mark=x, solid, mark options={scale=1, solid}] coordinates {
(48.6, 86.5)
(45.8, 86.6)
(43.0, 86.3)
(40.1, 86.2)
(37.8, 85.9)
(35.0, 86.0)
(33.1, 86.0)
(30.8, 85.4)
(28.9, 85.0)
(28.0, 83.3)
	};
        \addlegendentry{Transformer Sparsity = 0.5};

        \addplot[color={rgb,255:red,0;green,153;blue,0},mark=x, solid, mark options={scale=1, solid}] coordinates {
(43.8, 86.0)
(41.1, 85.9)
(38.0, 86.0)
(35.1, 86.0)
(32.7, 85.6)
(30.2, 85.3)
(27.9, 85.4)
(26.0, 85.1)
(24.4, 84.1)
(23.2, 83.3)
	};
        \addlegendentry{Transformer Sparsity = 0.6};

        \addplot[color=blue, mark=square, densely dotted, mark options={scale=1, solid}] coordinates {
(67.5, 87.0)
(61.9, 87.0)
(57.1, 87.1)
(46.9, 86.9)
(42.3, 86.8)
(33.1, 86.0)
(27.9, 85.4)
(26.0, 85.1)
	};
        \addlegendentry{Selected Points};
    
	\end{axis}
\end{tikzpicture}}
          \vskip -0.1in
          \caption{Intent Classification Accuracy (\%) on the SLURP test set.}
          \label{fig:slu-sep-size}
     \end{subfigure}
     \vskip -0.1in
     \caption{Model selection for \methodSS. As described in Sec.~\ref{subsec:joint-size}, we perform grid search over the Transformer sparsity (0.1 to 0.4/0.6) and CNN sparsity (0.1 to 0.95). The Pareto frontiers are shown in \textcolor{blue}{blue}, which are also presented in Fig.~\ref{fig:asr-main} and Fig.~\ref{fig:slu-main}.}
     \label{fig:model-select-sep-size}
    \vskip -0.15in
\end{figure}

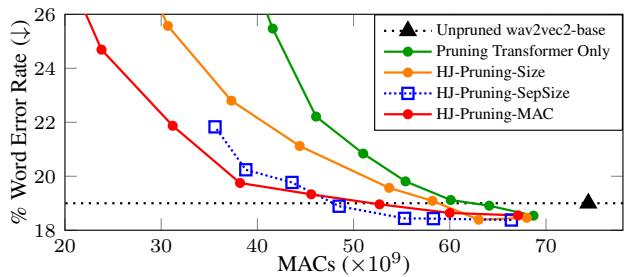
\begin{figure}[t]
\centering
\resizebox {0.98\linewidth} {!} {
\begin{tikzpicture}
	\begin{axis}[
		xlabel=MACs ($\times 10^9$),
		ylabel=\% Word Error Rate ($\downarrow$),
		xtick={0, 10,...,100},
    	xmin=20,
            xmax=78,
            ymin=18,
            ymax=26,
            ylabel shift = -5pt,
            xlabel shift = -6pt,
            label style={font=\footnotesize},
            ticklabel style={font=\scriptsize},
            width=\linewidth,
            height=0.5\linewidth,
	    every axis plot/.append style={thick},
            legend cell align={left},
            legend columns=1,
            legend style={at={(1,1)},anchor=north east,nodes={scale=0.65, transform shape}}
		]

        \addplot[color=black, dotted, mark=triangle*, mark options={scale=1.5, solid, fill=black}] coordinates {
(200, 19.00)
(74.4, 19.00)
(0, 19.00)
	};
        \addlegendentry{Unpruned wav2vec2-base};
        
	\addplot[color=green!60!black, solid, mark=*, mark options={scale=0.7, solid}] coordinates {
(68.7, 18.54)
(64.1, 18.91)
(60.1, 19.12)
(55.4, 19.81)
(51.0, 20.84)
(46.1, 22.21)
(41.6, 25.47)
(36.9, 29.81)
(32.3, 40.15)
	}; 
        \addlegendentry{Pruning Transformer Only};
        
	\addplot[color=orange, mark=*, solid, mark options={scale=0.7, solid}] coordinates {
(68.0, 18.46)
(63.0, 18.39)
(58.2, 19.09)
(53.7, 19.57)
(44.4, 21.12)
(37.3, 22.80)
(30.7, 25.57)
(24.0, 30.94)
(16.1, 44.67)
	};
        \addlegendentry{\methodOS};

        \addplot[color=blue, mark=square, densely dotted, mark options={scale=1, solid}] coordinates {
(66.4, 18.38)
(58.3, 18.43)
(55.3, 18.44)
(48.5, 18.89)
(43.6, 19.77)
(38.8, 20.24)
(35.6, 21.83)
	};
        \addlegendentry{\methodSS};
 
        \addplot[color=red, mark=*, solid, mark options={scale=0.7, solid, fill=red}] coordinates {
(67.1, 18.55)
(60.0, 18.64)
(52.7, 18.96)
(45.6, 19.33)
(38.2, 19.75)
(31.2, 21.87)
(23.8, 24.69)
(17.0, 30.20)
(9.9, 46.10)	
};
        \addlegendentry{\methodOM};
	\end{axis}
\end{tikzpicture}
} 
    \vskip -0.15in
  \caption{Robustness analysis. All models are trained on LibriSpeech 100h and then directly evaluated on the \textbf{out-of-domain Tedlium test set}. The trend is similar to that of the in-domain evaluation in Fig.~\ref{fig:asr-main}.}
  \label{fig:robustness}
  \vskip -0.15in
\end{figure}
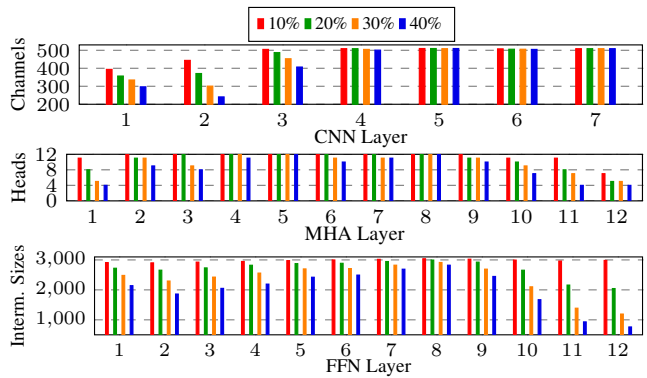
\begin{figure}
     \begin{subfigure}[b]{\linewidth}
          \centering
          \resizebox{\linewidth}{!}{\begin{tikzpicture}
    \begin{axis}[
            ybar,
            enlarge x limits=true,
            ymajorgrids=true,
            grid style={dashed,gray},
		xlabel=CNN Layer,
		ylabel=Channels,
		xtick=data,
            bar width=2pt,
            tickwidth=0pt,
            ymin=200,
            ymax=530,
            ylabel shift = -2pt,
            xlabel shift = -5pt,
            label style = {font=\scriptsize},
            ticklabel style={font=\scriptsize},
            width=\linewidth,
            height=0.27\linewidth,
            legend cell align={left},
            legend columns=-1,
            legend style={at={(0.3,1.05)},anchor=south west,nodes={scale=0.7, transform shape}},
            legend image code/.code={
        \draw [#1] (0pt,-1.5pt) rectangle (2pt,2pt); },
		]

        \addplot[color=red, fill=red] coordinates {
(1, 394)
(2, 444)
(3, 505)
(4, 509)
(5, 510)
(6, 508)
(7, 509)
	};
        \addlegendentry{10\%};

        \addplot[color=green!60!black, fill=green!60!black] coordinates {
(1, 357)
(2, 371)
(3, 487)
(4, 509)
(5, 510)
(6, 506)
(7, 509)
	};
        \addlegendentry{20\%};

        \addplot[color=orange, fill=orange] coordinates {
(1, 335)
(2, 301)
(3, 453)
(4, 505)
(5, 509)
(6, 506)
(7, 509)
	};
        \addlegendentry{30\%};

        \addplot[color=blue!90!black, fill=blue!90!black] coordinates {
(1, 297)
(2, 241)
(3, 407)
(4, 501)
(5, 510)
(6, 505)
(7, 509)
	};
        \addlegendentry{40\%};
        
	\end{axis}
 
\end{tikzpicture}}  
     \end{subfigure}
     
     \vskip -0.1in
     \begin{subfigure}[b]{\linewidth}
          \centering
          \resizebox{\linewidth}{!}{\begin{tikzpicture}
    \begin{axis}[
            ybar,
            enlarge x limits={abs=9pt},
            ymajorgrids=true,
            grid style={dashed,gray},
		xlabel=MHA Layer,
		ylabel=Heads,
		xtick=data,
            ytick={0,4,8,12},
            label style = {font=\scriptsize},
            ticklabel style={font=\scriptsize},
            bar width=1pt,
            tickwidth=0pt,
            ymin=0,
            ymax=12,
            ylabel shift = -3pt,
            xlabel shift = -5pt,
            width=\linewidth,
            height=0.25\linewidth,
            legend cell align={left},
            legend columns=-1,
            legend style={at={(0,1)},anchor=south west,nodes={scale=0.7, transform shape}},
            legend image code/.code={
        \draw [#1] (0pt,-1.5pt) rectangle (2pt,2pt); },
		]

        \addplot[color=red, fill=red] coordinates {
(1, 11)
(2, 12)
(3, 12)
(4, 12)
(5, 12)
(6, 12)
(7, 12)
(8, 12) 
(9, 12)
(10, 11)
(11, 11)
(12, 7)
	};
        \addlegendentry{10\%};

        \addplot[color=green!60!black, fill=green!60!black] coordinates {
(1, 8)
(2, 11) 
(3, 12)
(4, 12)
(5, 12)
(6, 12)
(7, 12)
(8, 12)
(9, 11) 
(10, 10) 
(11, 8) 
(12, 5)
	};
        \addlegendentry{20\%};

        \addplot[color=orange, fill=orange] coordinates {
(1, 5)
(2, 11)
(3, 9)
(4, 12)
(5, 12)
(6, 11)
(7, 11)
(8, 12)
(9, 11) 
(10, 9) 
(11, 7) 
(12, 5)
	};
        \addlegendentry{30\%};

        \addplot[color=blue!90!black, fill=blue!90!black] coordinates {
(1, 4)
(2, 9)
(3, 8)
(4, 11)
(5, 12)
(6, 10)
(7, 11)
(8, 12)
(9, 10)
(10, 7) 
(11, 4)
(12, 4)
	};
        \addlegendentry{40\%};

        \legend{};
        
	\end{axis}
 
\end{tikzpicture}}  
     \end{subfigure}
     
     \vskip -0.1in
     \begin{subfigure}[b]{\linewidth}
          \centering
          \resizebox{\linewidth}{!}{\begin{tikzpicture}
    \begin{axis}[
            ybar,
            enlarge x limits={abs=9pt},
            ymajorgrids=true,
            grid style={dashed,gray},
		xlabel=FFN Layer,
		ylabel=Interm. Sizes,
		xtick=data,
            ytick={1000,2000,3000},
            label style = {font=\scriptsize},
            ticklabel style={font=\scriptsize},
            bar width=1pt,
            tickwidth=0pt,
            ymin=500,
            ymax=3100,
            ylabel shift = -2pt,
            xlabel shift = -5pt,
            width=\linewidth,
            height=0.3\linewidth,
            legend cell align={left},
            legend columns=-1,
            legend style={at={(0,1)},anchor=south west,nodes={scale=0.7, transform shape}},
            legend image code/.code={
        \draw [#1] (0pt,-1.5pt) rectangle (2pt,2pt); },
		]

        \addplot[color=red, fill=red] coordinates {
(1, 2911)
(2, 2902)
(3, 2927)
(4, 2949)
(5, 2978)
(6, 2999)
(7, 3020)
(8, 3039)
(9, 3025)
(10,2995)
(11,2962)
(12,2983)
	};
        \addlegendentry{10\%};

        \addplot[color=green!60!black, fill=green!60!black] coordinates {
(1, 2723)
(2, 2658)
(3, 2734)
(4, 2823)
(5, 2876)
(6, 2888)
(7, 2947)
(8, 2996)
(9, 2927)
(10,2657)
(11,2161)
(12,2045)
	};
        \addlegendentry{20\%};

        \addplot[color=orange, fill=orange] coordinates {
(1, 2483)
(2, 2296)
(3, 2427)
(4, 2561)
(5, 2700)
(6, 2712)
(7, 2825)
(8, 2915)
(9, 2693)
(10,2102) 
(11,1389)
(12,1194)
	};
        \addlegendentry{30\%};

        \addplot[color=blue!90!black, fill=blue!90!black] coordinates {
(1, 2140)
(2, 1862)
(3, 2051)
(4, 2193)
(5, 2423)
(6, 2494)
(7, 2691)
(8, 2825)
(9, 2449)
(10,1672)
(11,932)
(12,763)
	};
        \addlegendentry{40\%};

        \legend{};
        
	\end{axis}
 
\end{tikzpicture}}  
     \end{subfigure}
     \vskip -0.15in
     \caption{ASR model architectures after \methodOM. The target sparsity ranges from 10\% to 40\%.}
     \label{fig:pruned-arch}
    \vskip -0.1in
\end{figure}

\begin{figure}[t]
\centering
\resizebox {0.95\linewidth} {!} {
\begin{tikzpicture}
	\begin{axis}[
		xlabel=MACs ($\times 10^9$),
		ylabel=\% Word Error Rate ($\downarrow$),
		xtick={0, 10,...,100},
            ytick={5.73,7,9,11,13},
    	xmin=14,
            xmax=78,
            ymin=5.2,
            ymax=13.7,
            ylabel shift = -5pt,
            xlabel shift = -5pt,
            label style={font=\footnotesize},
            ticklabel style={font=\footnotesize},
            width=\linewidth,
            height=0.45\linewidth,
	    every axis plot/.append style={thick},
            legend cell align={left},
            legend columns=1,
            legend style={at={(1,1)},anchor=north east,nodes={scale=0.7, transform shape}}
		]

        \addplot[color=black, dotted, mark=triangle*, mark options={scale=1.5, solid, fill=black}] coordinates {
(0, 5.73)
(74.4, 5.73)
(200, 5.73)
	};
        \addlegendentry{Unpruned HuBERT-base};

        \addplot[color=brown, dotted, mark=diamond*, mark options={scale=1.3, solid, fill=brown}] coordinates {
(34.92, 13.34)
	};
        \addlegendentry{DistilHuBERT$^*$};

        \addplot[color=green!60!black, dotted, mark=square*, mark options={scale=1, solid, fill=green!60!black}] coordinates {
(38.12, 8.33)
	};
        \addlegendentry{LightHuBERT-small$^*$};

        \addplot[color=cyan, dotted, mark=square*, mark options={scale=1, solid, fill=cyan}] coordinates {
(59.6, 6.72)
	};
        \addlegendentry{LightHuBERT-base$^*$};

        \addplot[color=red, mark=*, solid, mark options={scale=0.7, solid, fill=red}] coordinates {
(67.1, 5.52)
(59.9, 5.71)
(52.8, 5.72)
(45.6, 5.81)
(38.4, 6.05)
(31.3, 6.66)
(24.3, 8.02)
(16.8, 11.09)
(9.6, 23.49)	
};
        \addlegendentry{\methodOM};
	\end{axis}
\end{tikzpicture}
} 
    \vskip -0.15in
  \caption{Results of pruning HuBERT-base on LibriSpeech test-clean. $^*$ WERs from SUPERB~\cite{superb}. See Sec.~\ref{subsec:other-compression} for discussions.
  }
  \label{fig:hubert}
  \vskip -0.15in
\end{figure}
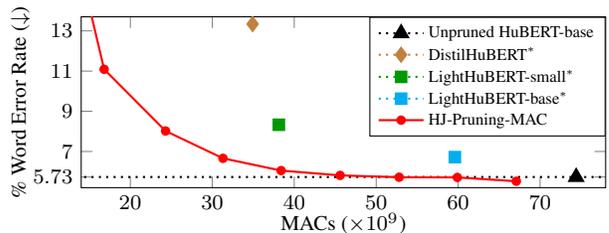

\subsection{Main results}
\label{subsec:main-results}
Fig.~\ref{fig:asr-main} compares various pruning methods for LibriSpeech ASR. The unpruned model has good performance (5.77\% WER) but is computationally expensive (74.4 GMACs). At a low sparsity ($>$55 GMACs), all pruned models achieve similar WERs which are even better than the original result, because the pruning target can regularize the training. As the sparsity increases, the baseline method which only prunes Transformer drastically degrades. Our proposed three algorithms which jointly prune CNN and Transformer consistently outperform the baseline by a large margin. We can reduce over 40\% of the total computation without degradation in WER.
\methodOM has similar performance with \methodSS, both outperforming \methodOS. This is because the CNN has much fewer parameters than Transformer. If we simply set an overall size sparsity, the pruned parameters are mainly from Transformer, while CNN still has high computational overhead. To prune them based on separate sizes (Eq.~\eqref{eq:prune-sep-size}), we have to search for the best combination of the two target sparsities. This model selection procedure is presented in Fig.~\ref{fig:asr-sep-size}, where we perform a grid search and select the Pareto frontiers. This requires much more computation than the other methods. Hence, the \methodOM is probably the best method in terms of performance and complexity. 

Fig.~\ref{fig:slu-main} shows the results of intent classification on SLURP. The overall trend is very similar to that of ASR. Our joint pruning methods outperform the baseline by a large margin, especially at a high sparsity (low MACs). \methodSS is comparable with \methodOM, but again, it requires a grid search over the two target sparsities as shown in Fig.~\ref{fig:slu-sep-size}. This high complexity hinders its usage in practice. Compared to ASR, we can achieve a higher compression rate (over 55\%) without loss in accuracy. This is probably because the classification task is easier and thus requires less information than the sequence-to-sequence task.

\subsection{Robustness of structured pruning}
\label{subsec:robustness}

To investigate the robustness of the proposed structured pruning methods, we test the ASR models using an \textit{out-of-domain} corpus, TED-LIUM~\cite{tedlium-corpus}. Note that these models are trained only with LibriSpeech data. As shown in Fig.~\ref{fig:robustness}, again, our joint pruning methods consistently outperform the baseline, and the trend is very similar to that of the in-domain evaluation (see Fig.~\ref{fig:asr-main}). This demonstrates that our pruning methods are robust.

\subsection{Architectures of pruned models}

Fig.~\ref{fig:pruned-arch} shows the remaining CNN channels, attention heads and FFN intermediate sizes after \methodOM. The target sparsity ranges from 10\% to 40\%. For CNN, the sequence length gradually decreases due to downsampling. The first few layers have higher computational cost, so they tend to be pruned more. For MHA and FFN, the upper layers are pruned the most, indicating that upper layers are more redundant. Prior studies had similar observations by analyzing the self-attention patterns in speech encoders~\cite{diagnality, branchformer, maekaku22interspeech}. The overall trend is also consistent with a prior work in NLP~\cite{cofi}.

\subsection{Comparison with other compression methods}
\label{subsec:other-compression}

As introduced in Sec.~\ref{subsec:ssl-models}, \method can be directly applied to other SSL models. In Fig.~\ref{fig:hubert}, we prune the HuBERT-base model based on the overall MACs for ASR. The performance is similar to that of the wav2vec2. We also include other compressed models for comparison, including DistilHuBERT~\cite{distilhubert} and LightHuBERT~\cite{lighthubert}. 
Note that these results are not really comparable due to: (1) Their WERs are from SUPERB~\cite{superb}, which combines a frozen SSL model with another learnable RNN. We also tried to replace the RNN with a single linear layer and fine-tune the entire model (same as our setting), but the performance was clearly worse. (2) Their compressed models are initially distilled using the 960h unlabeled LibriSpeech data and then fine-tuned on the 100h labeled data, but our task-specific pruning \textit{only} utilizes the 100h data. This comparison shows that our task-specific pruning method is highly effective.

\section{Conclusion}
In this paper, we propose \method to jointly prune heterogeneous components of SSL speech models, which achieves strong performance-efficiency tradeoffs compared to several baselines.
At a small sparsity (0.1 to 0.3), \method improves the wav2vec2 baseline while being faster.
Depending on the task, \method saves 40\% or 50\% MACs while maintaining the performance of wav2vec2.
\method is a general method that can be applied to most of speech SSL models such as HuBERT.
In the future, we plan to explore the application of \method on encoder-decoder SSL models~\cite{wu2022wav2seq} and other SLU tasks~\cite{lugosch2019speech,slue}.

\newpage

\let\OLDthebibliography\thebibliography
\renewcommand\thebibliography[1]{
  \OLDthebibliography{#1}
  \setlength{\parskip}{0.5pt}
  \setlength{\itemsep}{0.5pt plus 0.3ex}
}
\bibliographystyle{IEEEbib}
\bibliography{refs}

\begin{thebibliography}{10}

\bibitem{superb}
S.~w.~Yang, P.-H. Chi, Y.-S. Chuang, et~al.,
\newblock ``{SUPERB: Speech Processing Universal PERformance Benchmark},''
\newblock in {\em Proc. Interspeech}, 2021.

\bibitem{ssl-review}
A.~Mohamed, H.-y. Lee, L.~Borgholt, et~al.,
\newblock ``{Self-Supervised Speech Representation Learning: A Review},''
\newblock {\em arXiv:2205.10643}, 2022.

\bibitem{ssl-for-asr}
X.~Chang, T.~Maekaku, P.~Guo, et~al.,
\newblock ``An exploration of self-supervised pretrained representations for
  end-to-end speech recognition,''
\newblock in {\em Proc. ASRU}, 2021.

\bibitem{ssl-for-se-ss}
Z.~Huang, S.~Watanabe, S.-w. Yang, et~al.,
\newblock ``{Investigating Self-Supervised Learning for Speech Enhancement and
  Separation},''
\newblock in {\em Proc. ICASSP}, 2022.

\bibitem{slue}
S.~Shon, A.~Pasad, F.~Wu, et~al.,
\newblock ``{SLUE: New Benchmark Tasks For Spoken Language Understanding
  Evaluation on Natural Speech},''
\newblock in {\em Proc. ICASSP}, 2022.

\bibitem{espnet-slu}
S.~Arora, S.~Dalmia, P.~Denisov, et~al.,
\newblock ``{ESPnet-SLU: Advancing Spoken Language Understanding Through
  ESPnet},''
\newblock in {\em Proc. ICASSP}, 2022.

\bibitem{ssl-for-slu}
Y.~Peng, S.~Arora, Y.~Higuchi, et~al.,
\newblock ``{A Study on the Integration of Pre-trained SSL, ASR, LM and SLU
  Models for Spoken Language Understanding},''
\newblock in {\em Proc. SLT}, 2022.

\bibitem{wav2vec2}
A.~Baevski, Y.~Zhou, A.~Mohamed, and M.~Auli,
\newblock ``wav2vec 2.0: A framework for self-supervised learning of speech
  representations,''
\newblock in {\em Proc. NeurIPS}, 2020.

\bibitem{hubert}
W.-N. Hsu, B.~Bolte, Y.-H.~H. Tsai, et~al.,
\newblock ``{HuBERT: Self-supervised speech representation learning by masked
  prediction of hidden units},''
\newblock {\em IEEE/ACM Trans. Audio, Speech, Lang. Process.}, vol. 29, pp.
  3451--3460, 2021.

\bibitem{wavlm}
S.~Chen, C.~Wang, Z.~Chen, et~al.,
\newblock ``{WavLM: Large-scale self-supervised pre-training for full stack
  speech processing},''
\newblock {\em IEEE Journal of Selected Topics in Signal Processing}, 2022.

\bibitem{hinton2015distilling}
G.~Hinton, O.~Vinyals, J.~Dean, et~al.,
\newblock ``Distilling the knowledge in a neural network,''
\newblock {\em arXiv:1503.02531}, 2015.

\bibitem{distilbert}
V.~Sanh, L.~Debut, J.~Chaumond, and T.~Wolf,
\newblock ``{DistilBERT, a distilled version of BERT: smaller, faster, cheaper
  and lighter},''
\newblock {\em arXiv:1910.01108}, 2019.

\bibitem{tinybert}
X.~Jiao, Y.~Yin, et~al.,
\newblock ``{T}iny{BERT}: Distilling {BERT} for natural language
  understanding,''
\newblock in {\em Findings of EMNLP}, 2020.

\bibitem{shrinking-bigfoot}
Z.~Peng, A.~Budhkar, I.~Tuil, et~al.,
\newblock ``Shrinking bigfoot: Reducing wav2vec 2.0 footprint,''
\newblock in {\em SustaiNLP}, 2021.

\bibitem{distilhubert}
H.-J. Chang, S.-w. Yang, and H.-y. Lee,
\newblock ``{DistilHuBERT: Speech representation learning by layer-wise
  distillation of hidden-unit BERT},''
\newblock in {\em Proc. ICASSP}, 2022.

\bibitem{lighthubert}
R.~Wang, Q.~Bai, J.~Ao, et~al.,
\newblock ``{LightHuBERT: Lightweight and Configurable Speech Representation
  Learning with Once-for-All Hidden-Unit BERT},''
\newblock in {\em Proc. Interspeech}, 2022.

\bibitem{fithubert}
Y.~Lee, K.~Jang, J.~Goo, et~al.,
\newblock ``{FitHuBERT: Going Thinner and Deeper for Knowledge Distillation of
  Speech Self-Supervised Models},''
\newblock in {\em Proc. Interspeech}, 2022.

\bibitem{han2015pruning}
S.~Han, J.~Pool, et~al.,
\newblock ``Learning both weights and connections for efficient neural
  network,''
\newblock in {\em Proc. NeurIPS}, 2015.

\bibitem{filter-pruning-conv}
H.~Li, A.~Kadav, I.~Durdanovic, et~al.,
\newblock ``{Pruning Filters for Efficient ConvNets},''
\newblock in {\em Proc. ICLR}, 2017.

\bibitem{network-slimming}
Z.~Liu, J.~Li, Z.~Shen, et~al.,
\newblock ``{Learning Efficient Convolutional Networks Through Network
  Slimming},''
\newblock in {\em Proc. ICCV}, 2017.

\bibitem{platon-icml22}
Q.~Zhang, S.~Zuo, C.~Liang, et~al.,
\newblock ``{PLATON}: Pruning large transformer models with upper confidence
  bound of weight importance,''
\newblock in {\em Proc. ICML}, 2022.

\bibitem{asapp-pruning}
Z.~Wang, J.~Wohlwend, and T.~Lei,
\newblock ``{Structured Pruning of Large Language Models},''
\newblock in {\em Proc. EMNLP}, 2020.

\bibitem{cofi}
M.~Xia, Z.~Zhong, and D.~Chen,
\newblock ``{Structured Pruning Learns Compact and Accurate Models},''
\newblock in {\em Proc. ACL}, 2022.

\bibitem{super-tickets-nlp}
C.~Liang, S.~Zuo, M.~Chen, et~al.,
\newblock ``Super tickets in pre-trained language models: From model
  compression to improving generalization,''
\newblock in {\em Proc. ACL}, 2021.

\bibitem{structured-prune-rnn}
P.~Dong, S.~Wang, W.~Niu, et~al.,
\newblock ``Rtmobile: Beyond real-time mobile acceleration of rnns for speech
  recognition,''
\newblock in {\em ACM/IEEE Design Automation Conference (DAC)}, 2020.

\bibitem{prune-lstm}
S.~Wang, P.~Lin, R.~Hu, et~al.,
\newblock ``{Acceleration of LSTM With Structured Pruning Method on FPGA},''
\newblock {\em IEEE Access}, 2019.

\bibitem{deliang-compressing-enhancement}
K.~Tan and D.L. Wang,
\newblock ``{Compressing Deep Neural Networks for Efficient Speech
  Enhancement},''
\newblock in {\em Proc. ICASSP}, 2021.

\bibitem{group-sparse-reg}
S.~Scardapane, D.~Comminiello, A.~Hussain, and A.~Uncini,
\newblock ``Group sparse regularization for deep neural networks,''
\newblock {\em Neurocomputing}, vol. 241, pp. 81--89, 2017.

\bibitem{parp}
C.-I~J. Lai, Y.~Zhang, A.~H. Liu, et~al.,
\newblock ``{PARP: Prune, Adjust and Re-Prune for Self-Supervised Speech
  Recognition},''
\newblock in {\em Proc. NeurIPS}, 2021.

\bibitem{felix-sew}
F.~Wu, K.~Kim, J.~Pan, et~al.,
\newblock ``{Performance-Efficiency Trade-offs in Unsupervised Pre-training for
  Speech Recognition},''
\newblock in {\em Proc. ICASSP}, 2022.

\bibitem{librispeech-corpus}
V.~Panayotov, G.~Chen, D.~Povey, and S.~Khudanpur,
\newblock ``{Librispeech: An ASR corpus based on public domain audio books},''
\newblock in {\em Proc. ICASSP}, 2015.

\bibitem{transformer}
A.~Vaswani, N.~Shazeer, N.~Parmar, et~al.,
\newblock ``Attention is all you need,''
\newblock in {\em Proc. NeurIPS}, 2017.

\bibitem{gelu}
D.~Hendrycks and K.~Gimpel,
\newblock ``{Gaussian Error Linear Units (GELUs)},''
\newblock {\em arXiv:1606.08415}, 2016.

\bibitem{l0sparse-iclr18}
C.~Louizos, M.~Welling, and D.~P. Kingma,
\newblock ``{Learning Sparse Neural Networks through L0 Regularization},''
\newblock in {\em ICLR}, 2018.

\bibitem{pytorch}
A.~Paszke et~al.,
\newblock ``Pytorch: An imperative style, high-performance deep learning
  library,''
\newblock {\em Proc. NeurIPS}, 2019.

\bibitem{huggingface-transformers}
T.~Wolf et~al.,
\newblock ``Huggingface's transformers: State-of-the-art natural language
  processing,''
\newblock {\em arXiv:1910.03771}, 2019.

\bibitem{tedlium-corpus}
A.~Rousseau et~al.,
\newblock ``{TED-LIUM}: an automatic speech recognition dedicated corpus.,''
\newblock in {\em Proc. LREC}, 2012.

\bibitem{ctc}
A.~Graves, S.~Fern{\'a}ndez, F.~Gomez, et~al.,
\newblock ``Connectionist temporal classification: labelling unsegmented
  sequence data with recurrent neural networks,''
\newblock in {\em Proc. ICML}, 2006.

\bibitem{slurp-corpus}
E.~Bastianelli, A.~Vanzo, P.~Swietojanski, and V.~Rieser,
\newblock ``{SLURP: A Spoken Language Understanding Resource Package},''
\newblock in {\em {Proc. EMNLP}}, 2020.

\bibitem{diagnality}
S.~Zhang, E.~Loweimi, P.~Bell, and S.~Renals,
\newblock ``On the usefulness of self-attention for automatic speech
  recognition with transformers,''
\newblock in {\em Proc. SLT}, 2021.

\bibitem{branchformer}
Y.~Peng, S.~Dalmia, I.~Lane, and S.~Watanabe,
\newblock ``Branchformer: Parallel {MLP}-attention architectures to capture
  local and global context for speech recognition and understanding,''
\newblock in {\em Proc. ICML}, 2022.

\bibitem{maekaku22interspeech}
T.~Maekaku, Y.~Fujita, Y.~Peng, and S.~Watanabe,
\newblock ``{Attention Weight Smoothing Using Prior Distributions for
  Transformer-Based End-to-End ASR},''
\newblock in {\em Proc. Interspeech}, 2022.

\bibitem{wu2022wav2seq}
F.~Wu, K.~Kim, S.~Watanabe, et~al.,
\newblock ``Wav2seq: Pre-training speech-to-text encoder-decoder models using
  pseudo languages,''
\newblock {\em arXiv:2205.01086}, 2022.

\bibitem{lugosch2019speech}
L.~Lugosch, M.~Ravanelli, P.~Ignoto, et~al.,
\newblock ``Speech model pre-training for end-to-end spoken language
  understanding,''
\newblock in {\em Interspeech}, 2019.

\end{thebibliography}

\end{document}